          [2001/04/25 1.1 (PWD)]
\documentclass[a4paper,twocolumn]{esapub2005} % European paper
\usepackage{times}
\usepackage[numbers,sort&compress]{natbib}
\usepackage{url}
\usepackage{graphicx}
\usepackage{booktabs}
\usepackage{subcaption}
\usepackage{mathtools}
\usepackage{amsmath}
\usepackage{multirow}
\usepackage{soul}
\usepackage[dvipsnames]{xcolor}
\usepackage[compact]{titlesec}         % you need this package
\titlespacing{\section}{0pt}{0pt}{0pt} % this reduces space between (sub)sections to 0pt, for example
\usepackage{float}

\usepackage{hyperref}
\hypersetup{
    colorlinks=true,
    linkcolor=blue,
    filecolor=magenta,      
    urlcolor=cyan,
    pdftitle={Overleaf Example},
    pdfpagemode=FullScreen,
    }
\hypersetup{hidelinks}

\title{RANS: Highly-Parallelised Simulator for Reinforcement learning based Autonomous Navigating Spacecrafts}
\author{Matteo El-Hariry\footnote{Equal contribution.} \thanks{Presenting author.} , Antoine Richard$^*$, Miguel Olivares-Mendez}
\affil{\normalsize Space Robotics (SpaceR) Research Group, SnT-University of Luxembourg, \\ \normalsize Campus Kirchberg 29, Avenue John F. Kennedy L-1855 Luxembourg}
\affil{\footnotesize \{matteo.elhariry, antoine.richard, miguel.olivaresmendez\}@uni.lu}

\begin{document}
\maketitle

\begin{abstract}
Nowadays, realistic simulation environments are essential to validate and build reliable robotic solutions.
This is particularly true when using Reinforcement Learning (RL) based control policies.
To this end, both robotics and RL developers need tools and workflows to create physically accurate simulations and synthetic datasets.
Gazebo, MuJoCo, Webots, Pybullets or Isaac Sym are some of the many tools available to simulate robotic systems.
Developing learning-based methods for space navigation is, due to the highly complex nature of the problem, an intensive data-driven process that requires highly parallelized simulations.
When it comes to the control of spacecrafts, there is no easy to use simulation library designed for RL. 
%Hence, simulating spacecrafts, or their earth-based analog system: air-floating platforms, can be tedious.
We address this gap by harnessing the capabilities of NVIDIA Isaac Gym, where both physics simulation and the policy training reside on GPU.
%Building on this tool, we provide an open-source library enabling users to simulate thousands of parallel spacecrafts.
%With it, we provide different tasks, such as position and attitude control, but also velocity control. 
Building on this tool, we provide an open-source library enabling users to simulate thousands of parallel spacecrafts, that learn a set of maneuvering tasks, such as position, attitude, and velocity control. 
These tasks enable to validate complex space scenarios, such as trajectory optimization for landing, docking, rendezvous and more.

\end{abstract}

\section{Introduction}

\begin{figure*}[!bht]
    \centering
    \begin{tabular}{cc}
        \includegraphics[width=0.4\linewidth]{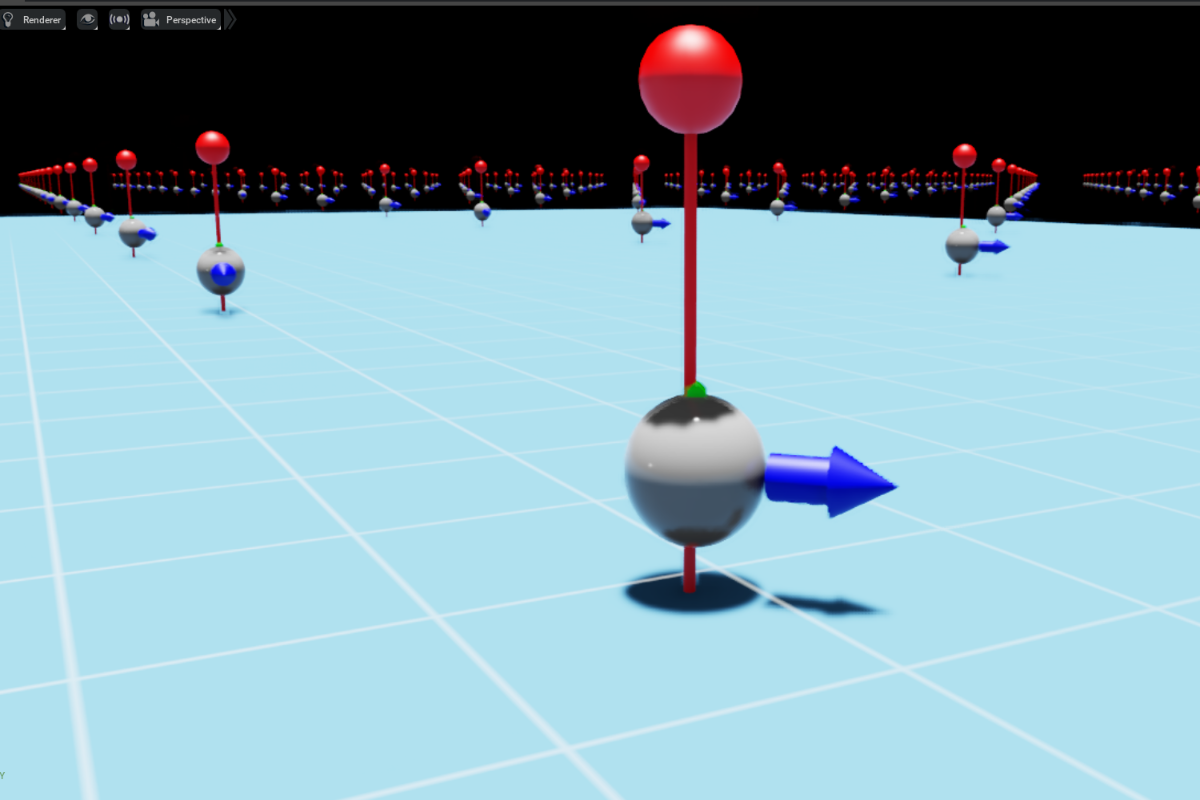} &\includegraphics[width=0.4\linewidth]{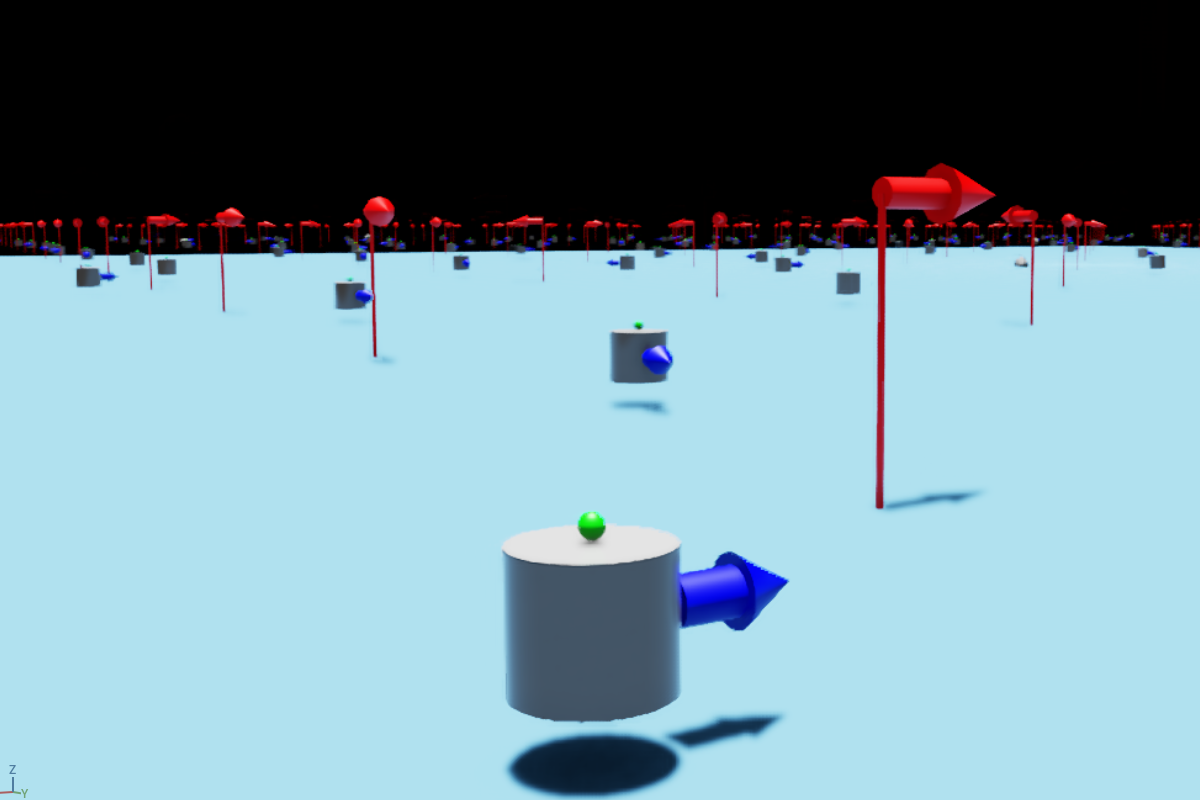}\\\includegraphics[width=0.4\linewidth]{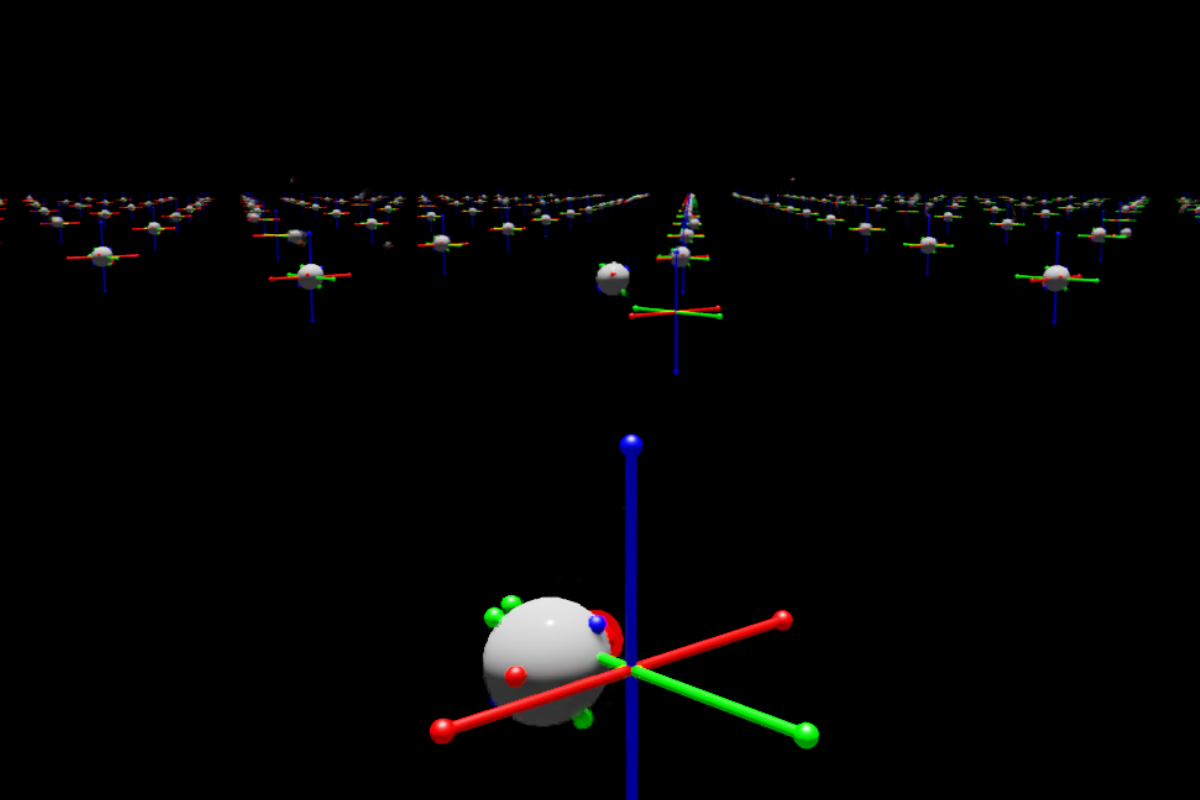}&\includegraphics[width=0.4\linewidth]{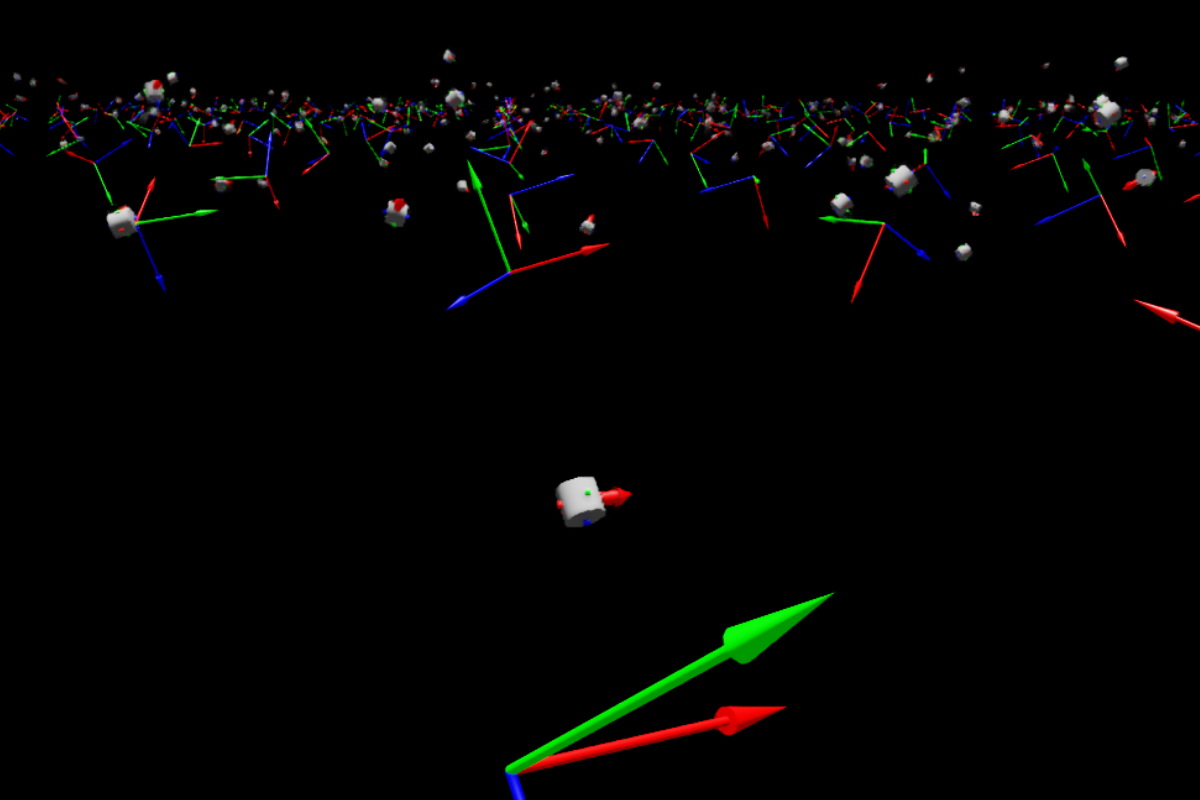}\\
         & 
    \end{tabular}
    \caption{Sample renders of both 3DoF and 6DoF environments when the UI is enabled. The Floating platform or satellites are represented as either a sphere or a cylinder, while the pin, and arrows indicate the goal the agent must reach.}
    \label{fig:render_examples}
\end{figure*}

The pursuit of space exploration has always been at the forefront of scientific and technological endeavors. As we venture further into the outer space, the challenges related to spacecraft missions become increasingly intricate. A central concern in the realm of space exploration is the design of robust and reliable guidance, navigation, and control (GNC) systems for onboard spacecraft. The effectiveness of these systems can significantly impact mission success and safety. Historically, missions have relied on conventional GNC techniques. However, as we progress into the future, the quest for enhanced mission performance, autonomy, and adaptability is pushing the boundaries of onboard GNC systems. Autonomous systems offer advantages in terms of control accuracy, mission flexibility, and adaptability, enabling swift responses to unforeseen contingencies~\cite{mauro2018survey_gnc}. Furthermore, they are essential for executing maneuvers in distant regions of space, where communication delays with Earth are substantial, and operational costs need to be minimized. The pursuit of autonomy in space missions introduces challenges ranging from improving onboard sensing and actuation capabilities to enhancing maneuver planning, mission management, and fault detection, isolation, and recovery (FDIR) mechanisms. Autonomous spacecraft must operate in an environment where human intervention is limited, demanding sophisticated solutions that can handle complex scenarios with precision and reliability. Moreover, the landscape of space missions is evolving with the emergence of small satellites, particularly nano- and micro-satellites, due to their cost-effectiveness and suitability for various Earth observation and scientific purposes. However, the downsizing of spacecraft places additional constraints on GNC systems. Limited onboard resources, including thrust, power, and computational capabilities, necessitate innovative GNC solutions tailored to these compact platforms.
\newline\newline
Simultaneously, there has been a significant increase in the adoption of Artificial Intelligence (AI), Deep Learning (DL) and Reinforcement Learning solutions (RL) for terrestrial robotics and even for space navigation systems~\cite{song2022dl_relative_navig}. These approaches, can greatly benefit autonomous spacecraft navigation by leveraging onboard sensors, such as cameras, for rendezvous, terrain navigation, and asteroid exploration tasks. Among the possible approaches, RL solutions have proven particularly adapt to tackle space GNC problems. Their application encompass a wide range of tasks, including planetary landing~\cite{gaudet2020deep}, path planning for lunar or asteroid hopping rovers~\cite{yu2021path_planning, tanaka2021hopper}, spacecraft orbit control within unknown gravitational fields~\cite{Willis2016ReinforcementLF}, and spacecraft map generation during orbits around small celestial bodies~\cite{chan2019mapping}. 
\newline\newline
However, the integration of RL-based solutions for space missions poses its own set of challenges. These solutions require thorough validation and testing before being deployed on the real systems, but they also need the right software framework within to be developed. Numerous simulation~\cite{korber2021comparing} tools such as Gazebo~\cite{Gazebo}, MuJoCo~\cite{todorov2012mujoco}, Webots~\cite{michel2004webots}, Pybullets~\cite{coumans2016pybullet}, and Isaac Sim are available, each with its own set of advantages and limitations in simulating robotic systems. When it comes to the control of spacecrafts, there is no easy to use simulation library designed for RL. 
\newline
This paper addresses the need for a highly parallelized simulator tailored specifically for RL-based autonomous spacecraft navigation. We introduce RANS (Reinforcement learning based Autonomous Navigating Spacecraft Simulator), an open-source Python package designed to simplify the training of RL agents for thrust-based spacecraft control. By harnessing the power of NVIDIA Isaac Gym~\cite{makoviychuk2021isaac}, RANS provides an environment that facilitates parallelized simulations crucial for learning-based methods in space navigation. The proposed simulator, is compatible with other popular RL frameworks, such as rl-games~\cite{rl-games2021}, for which we provide default agents, and demonstration models. It offers a user-friendly GUI, illustrated in Figure~\ref{fig:render_examples}, to visualize agent behavior across tasks. Rendering is available during both training and evaluation. The subsequent sections present the related works and elaborate on the architecture and results of RANS, demonstrating its potential in advancing research towards fully autonomous GNC systems for space exploration.

\section{Related Work}
\begin{figure*}[!thb] 
    \centering
    
    \includegraphics[width=\textwidth]{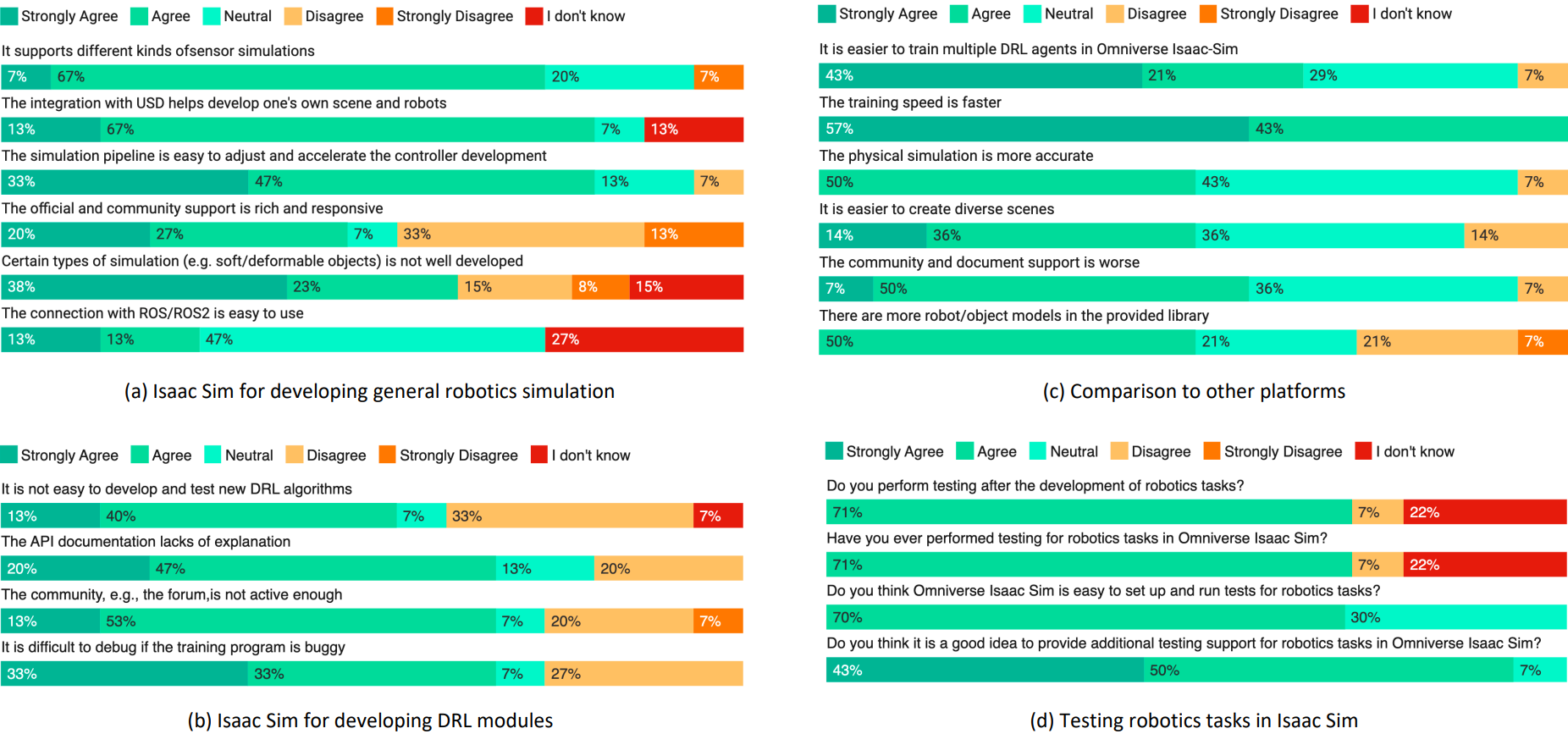}
    \caption{Survey on the advantages and limitations of Isaac Sim (source: Zhou et al.~\cite{zhou2023isaac_benchmark})}
    \label{fig:survey}
\end{figure*}

%\subsection{Robotics and RL}
In the realm of Deep Reinforcement Learning (DRL) for robotics, Isaac Sim stands out as a valuable simulation platform. A recent survey (Fig. \ref{fig:survey}), conducted by Zhou et al.~\cite{zhou2023isaac_benchmark}, highlights its strengths: diverse sensor support (74\%) and ease of creating custom scenes and robots using USD (Universal Scene Design) format (80\%). Although it excels in training speed (63\%) due to NVIDIA GPU compatibility, practitioners noted challenges in API documentation (67\%) and community support (66\%). Notably, practitioners find Isaac Sim favorable for training multiple DRL agents and emphasize the need for enhanced testing support (93\%). This survey underscores Isaac Sim's potential for DRL applications, while also pinpointing areas for improvement.

Despite its recent increased adoption and its positive evaluation from the community, a limited number of works that expanded Isaac Sim to create specific RL training environments for new robots or scenarios is available. Notable examples include ORBIT~\cite{mittal2023orbit}, which provides a suite of benchmark tasks encompassing various difficulty levels, from single-stage actions like cabinet opening and cloth folding to more complex multi-stage tasks such as room reorganization. Additionally, Zhou et al.~\cite{zhou2023isaac_benchmark} proposed a public industrial benchmark for robotic manipulation, augmenting Isaac Sim to facilitate benchmarking in this domain. Rojas et al.~\cite{rojas2022isaac_mobile_robots} created a simulator using Isaac Sim to train mobile robots with deep reinforcement learning. 

However, in the context of RL-driven navigation in space for free-flying objects, to the best of the authors' knowledge, no simulator has been created or publicly disclosed to date. The introduction of RANS attempts to mitigate this gap, offering a new alternative to design autonomous trajectories in 2D and 3D space.

%\subsection{Spacecrafts navigation}
The proposed framework, RANS, stands distinct from existing trajectory design and optimization tools specialized in space mission design. Tools like ORDEM \cite{ordem}, enabling flux analysis based on debris size and year, and Trajectory Browser \cite{trajectory_browser} for trajectory design to planets and small-bodies, operate within specific mission-focused paradigms. General Mission Analysis Tool (GMAT) \cite{gmat} offers open-source multi-mission support for space mission design and navigation. SPICE (Spacecraft Planet Instrument Camera-matrix Events) \cite{spice} supports CubeSat and smallsat missions, focusing on trajectory planning and science data analysis. In contrast, RANS is designed to facilitate the training of Reinforcement Learning (RL) agents for generic thrust-based spacecraft control, presenting a flexible approach with low level access to control actions and state variables.

\section{RANS: Overview and Architecture}
The primary objective of RANS is to bridge the gap between available simulation tools and the specialized requirements of RL-based spacecraft navigation.

RANS is structured to replicate realistic orbital operations as well as air-bearing platforms, providing a fast, stable, and precise simulation environment. It comprises two main scenarios: a 3 Degree of Freedom (DoF) "Floating Platform" (FP) robot and a 6 DoF navigating scenario. These scenarios enable users to specify or randomize initial conditions and goals for spacecraft control tasks.

The RANS architecture is specifically engineered to optimize simulation efficiency and parallelizability. By harnessing the advanced capabilities of NVIDIA OmniIsaacGym~\cite{makoviychuk2021isaac}, both physics simulation and policy training are conducted on the GPU, allowing for highly parallelized simulations. Remarkably, this approach facilitates running as many as 16,000 environments concurrently on a single GeForce RTX 4090 GPU, demonstrating a significant advancement in simulation throughput and computational performance.
\begin{table*}[!htb]
    \centering
    \resizebox{\linewidth}{!}{
	\begin{tabular}{ccccccccccc}
		\hline
        \textbf{Task}            & \textbf{$\text{tf}$} & \textbf{$\text{td}_1$} & \textbf{$\text{td}_2$} & \textbf{$\text{td}_3$}  & \textbf{$\text{td}_4$ } & \textbf{$\text{td}_5$ } & \textbf{$\text{td}_6$ } & \textbf{$\text{td}_7$ } & \textbf{$\text{td}_8$ } & \textbf{$\text{td}_9$ }\\ \hline
        3DoF Go to position  & 0           & $\Delta x$    & $\Delta y$    & -                     & -                     & & & & &\\
        3DoF Go to pose      & 1           & $\Delta x$    & $\Delta y$    & $\cos(\Delta \theta)$ & $\sin(\Delta \theta)$ & & & & &\\
        3DoF Track velocity  & 2           & $\Delta v_x$  & $\Delta v_y$  & -                     & -                     & & & & &\\ \hline
        6DoF Go to position  & 0           & $\Delta x$    & $\Delta y$    & $\Delta z$            & -                     & - & - & - & -&-\\
        6DoF Go to pose      & 1           & $\Delta x$    & $\Delta y$    & $\Delta z$            & $\Delta R[0,0]$       & $\Delta R[0,1]$ & $\Delta R[0,2]$&$\Delta R[1,0]$&$\Delta R[1,1]$ & $\Delta R[1,2]$\\
        6DoF Track velocity  & 2           & $\Delta v_x$  & $\Delta v_y$  & $\Delta v_z$          & -                     & - & - & - & - &  -\\ \hline
	\end{tabular}
}
\caption{State task-specific data for the 3DoF and 6DoF tasks.}
\label{tab:task_data}
\end{table*}

\subsection{Simulation engine}

%Describe the physics parameters and basic usage (dt, gravity, etc).

To replicate the environmental conditions, we utilize the PhysX engine within IsaacSim, a GPU-based physics engine renowned for its capacity to rapidly simulate numerous parallel systems. Although GPU-based physics engines may not offer the same level of precision as their CPU-based counterparts, they present an advantage in efficiency, enabling swift simulations, especially conducive for reinforcement learning tasks characterized by short time intervals. The inherent imprecision associated with GPU-based physics engines may not prominently affect these tasks. Furthermore, this imprecision can introduce a controlled level of noise into the system's state, potentially enhancing the resilience and adaptability of the learned agents.
% To simulate the environment, we rely on IsaacSim's PhysX engine. A GPU-based physics engine that allows to quickly simulate thousands of parallel systems. As a GPU-based physics engine, it is not as accurate as a CPU-based physics engine. Yet, for reinforcement learning tasks, which usually spawn short time periods, these inaccuracies are not obvious. Moreover, the less accurate physics can also introduce noise in the state of the system, potentially leading to improved robustness of the learned agents.

To make our simulation as stable as possible, we use a sub-stepping strategy. This amounts to step the physics engine faster than the controllers, or agents, are running such that the simulation remains stable. In practice, our agents actuate at 5 or 10 Hz, while the simulation is updated at 50 or 100 Hz respectively.

To simulate a floating platform, we disable the gravity from the simulator and only apply forces in the xy plane aligning the point of application of the force with the z position of its center of mass, in doing so, the system is bound to remain 3~DoF. Aside from this, we do not impose any other constraints. 
For the 6~DoF satellites, we do not impose any of these constraints such that the satellites can move freely along any direction.
\newline
In OmniIsaacGym, the physics engine does not provide automatic force composition, i.e. if a first force $f_1$ is applied on a body, and a second one $f_2$ is then applied, only $f_2$ is applied.
To solve this issue, we create as many rigid bodies at the center of the platform as there are thrusters, and apply forces at the relative position of the thrusters, this approach circumvents the current limitation in OmniIsaacGym.

\subsection{Environment and Tasks}

\begin{figure}
    \centering
    \begin{tabular}{cc}
        \includegraphics[width=0.4\linewidth]{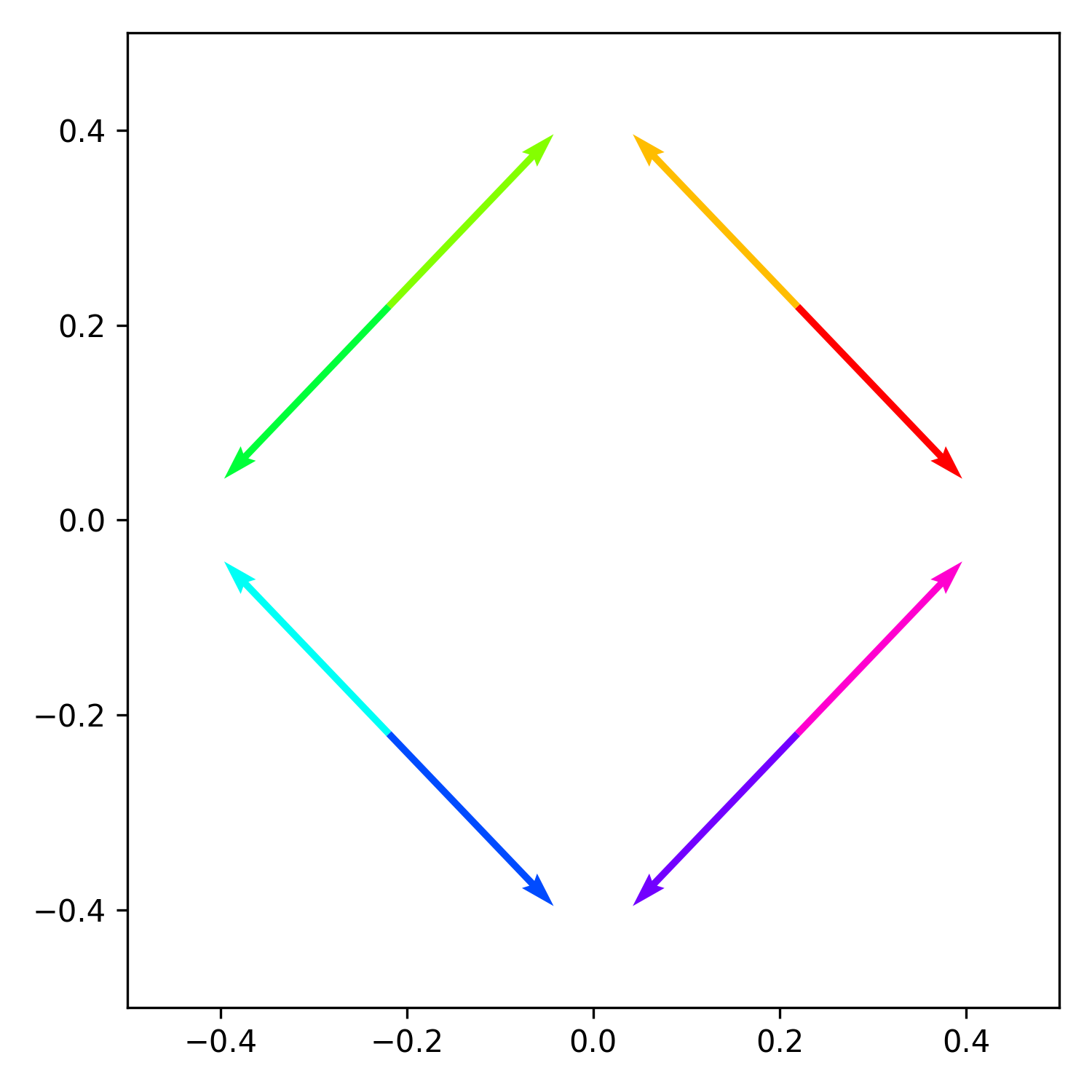} &  \includegraphics[width=0.4\linewidth]{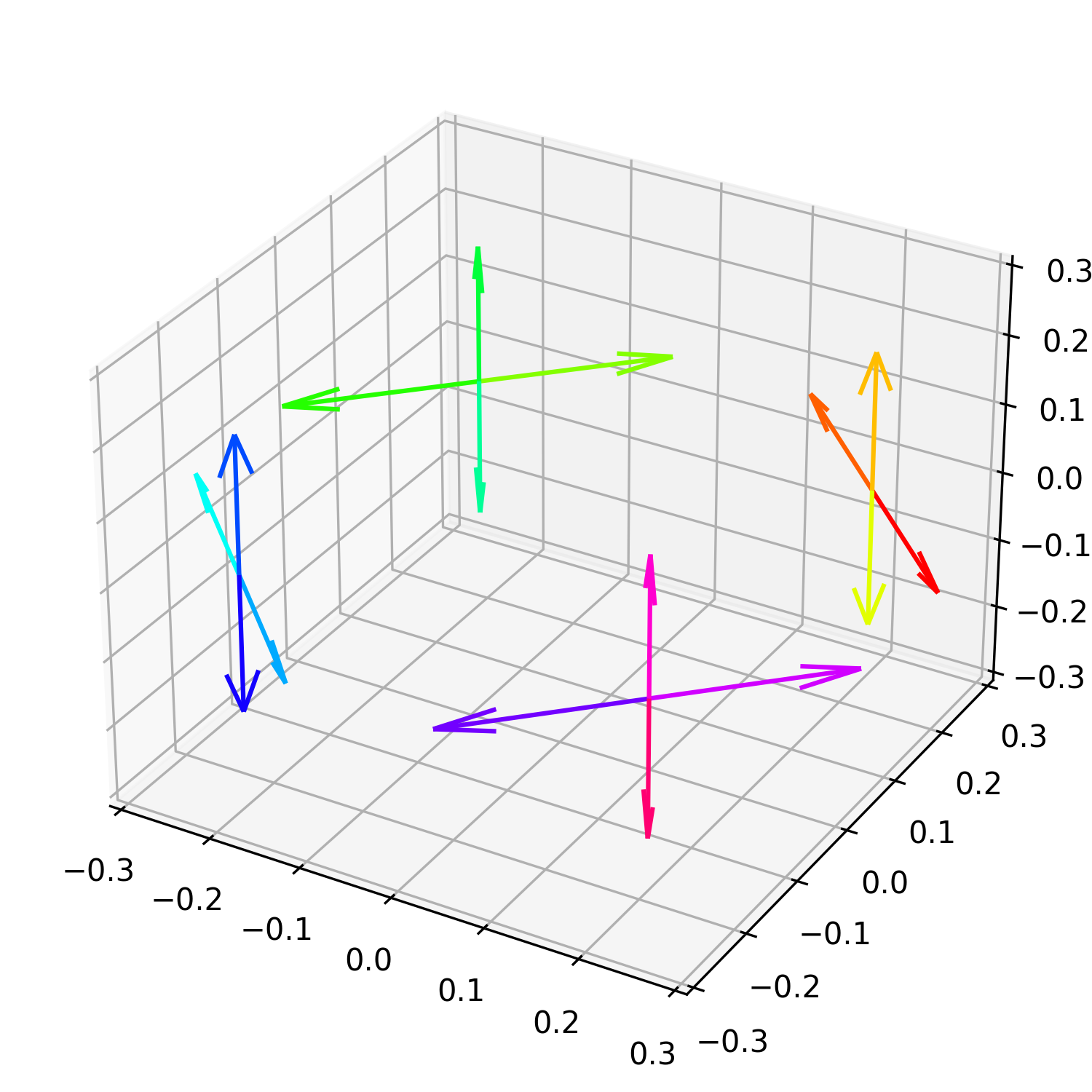}\\
        (a) 3DoF & (b) 6DoF 
    \end{tabular}
    
    \caption{The arrows indicate the direction of the forces applied by the thrusters mounted on the systems. the center of mass is located in (0,0,0)}
    \label{fig:thrusters_config}
\end{figure}

In the 3 DoF scenarios, the simulator includes a default system configuration with 8 thrusters (Fig.~\ref{fig:thrusters_config} (a)) and allows users to customize various parameters, such as mass and thruster positions, via configuration files. The tasks defined for position control and position-attitude control are named GoToXY and GoToPose-2D, respectively.
We also provide tasks for velocity control called TrackXYVelocity and TrackXYOVelocity. In the first one, the agent learns to track linear velocities in the xy plane, while in the latter, the agent learns to track both linear and angular velocities.
Similarly, in the 6 DoF scenario, the simulators comes with a default 16 thrusters configuration (Fig.~\ref{fig:thrusters_config} (b)) and a similar set of tasks. GoToXYZ and GoToPose-3D are defined to encompass a broader range of navigation challenges, enabling simulations and training scenarios that demand a higher level of control and precision in spacecraft movement and positioning. This extension involved augmenting the simulator's capabilities and configurations to accurately propagate the system's state in 3D while enabling the agent's movement along the Z axis. To estimate the precise orientation in 3D, we use a continuous 6D representation, that, unlike quaternions and Euler angles, do not suffer from discontinuities~\cite{zhou2019continuity}. This is particularly important as neural networks can struggle to map distances properly in Euler or quaternion spaces. To enable free-flying maneuvers in space, 8 additional thrusters aligned with the vertical axis were added at the same location as the previous ones. A visual representation of the thrusters configuration in the 3~DoF and 6~DoF cases is shown in Figure~\ref{fig:thrusters_config}.

Overall, for all tasks on the 3DoF scenarios, we use an observation space of size 10, $(\cos(\theta),\sin(\theta),v_{xy},\omega_z,\text{tf},\text{td}_{1-4})$,and an action space of size 8.
For the 6DoF scenarios, we use an observation space of size 22, $(6D, v_{xyz},\omega_{xyz},\text{tf},\text{td}_{1-9})$, and an action space of size 16.
With $v$ and $\omega$ the linear and angular respectively velocities along a given axis, $\theta$ the heading of the platform, $6D$ the orientation of the system casted as a 6D representation. $tf$ is a flag indicating the kind of task the agent should solve, 0 for GoToXY (or GoToXYZ), 1 for GoToPose, and 2 for TrackXYVelocities (or TrackXYZVelocities). 
Finally, $td$ contains the information required to solve the different tasks.

The task-specific data ($\text{td}$) are shown in Table~\ref{tab:task_data}, where $\Delta$ denotes the distance between each variable (position, velocity or angle) and their target value.
The observation space is structured as such so that this work can easily be extended to support multitask-capable policies. For the 6DoF GoToPose, $\Delta R$ is the rotation matrix between the rotation matrix of the system in the global frame $R_s$, and the goal rotation matrix in the global frame $R_g$ with $\Delta R = R_s^T R_g$.

\begin{figure*}[!ht]
   \centering
   \begin{subfigure}[b]{0.243\textwidth}
       \centering
       \includegraphics[width=\linewidth]{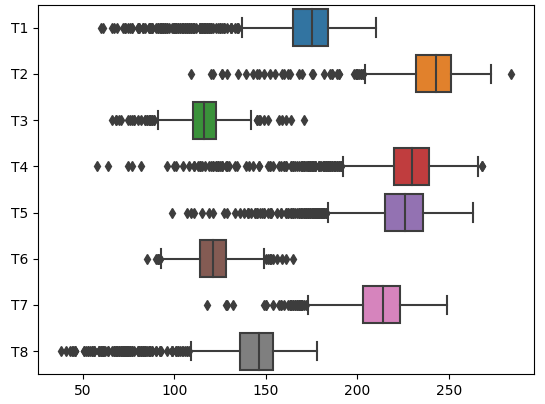}
       \caption{Actions count}
   \end{subfigure}
   \begin{subfigure}[b]{0.24\textwidth}
       \centering
       \includegraphics[width=\linewidth]{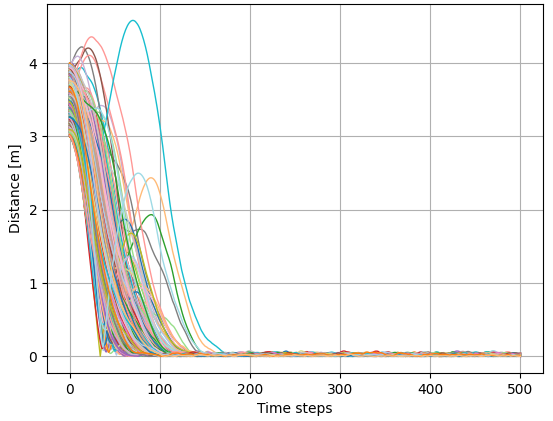}
       \caption{Position distances}
   \end{subfigure}
   \begin{subfigure}[b]{0.24\textwidth}
       \centering
       \includegraphics[width=\linewidth]{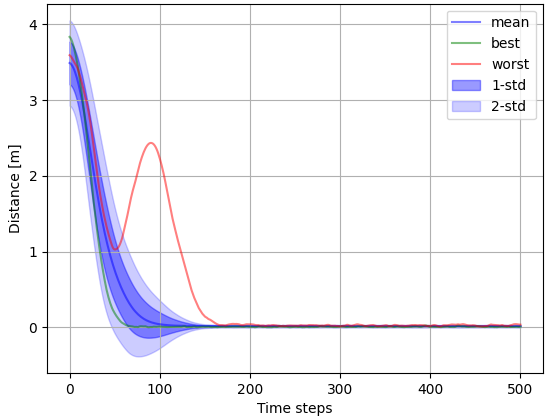}
       \caption{Position distances summary}
   \end{subfigure}
   \begin{subfigure}[b]{0.255\textwidth}
       \centering
       \includegraphics[width=\linewidth]{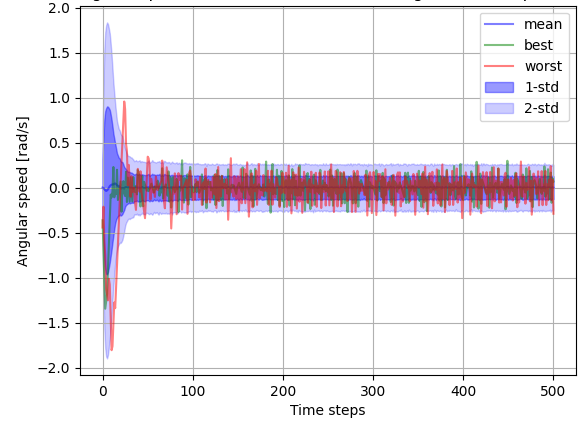}
       \caption{Angular velocities summary}
   \end{subfigure}
   
   \vspace{0.5cm} % Add vertical space between rows
       
   \begin{subfigure}[b]{0.24\textwidth}
       \centering
       \includegraphics[width=\linewidth]{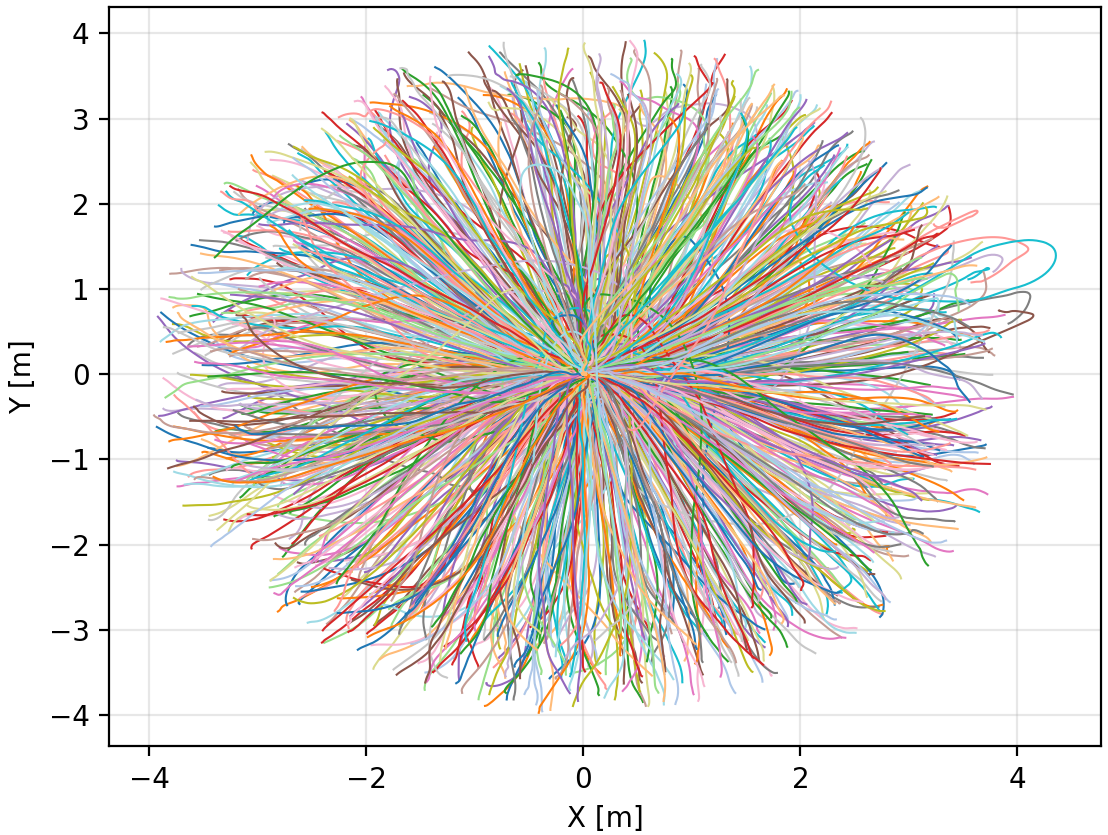}
       \caption{Trajectories}
   \end{subfigure}
   \begin{subfigure}[b]{0.24\textwidth}
       \centering
       \includegraphics[width=\linewidth]{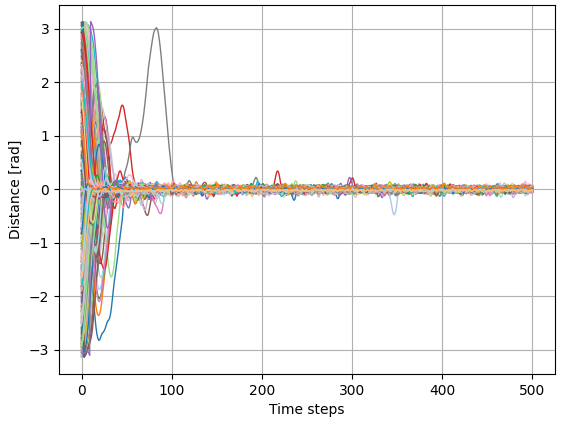}
       \caption{Angle distances}
   \end{subfigure}
   \begin{subfigure}[b]{0.24\textwidth}
       \centering
       \includegraphics[width=\linewidth]{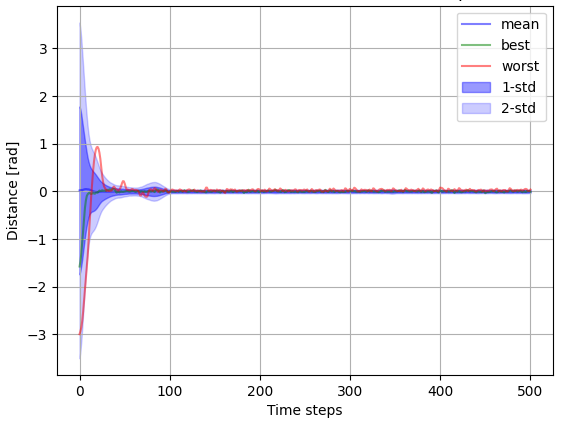}
       \caption{Angle distances summary}
   \end{subfigure}
   \begin{subfigure}[b]{0.245\textwidth}
       \centering
       \includegraphics[width=\linewidth]{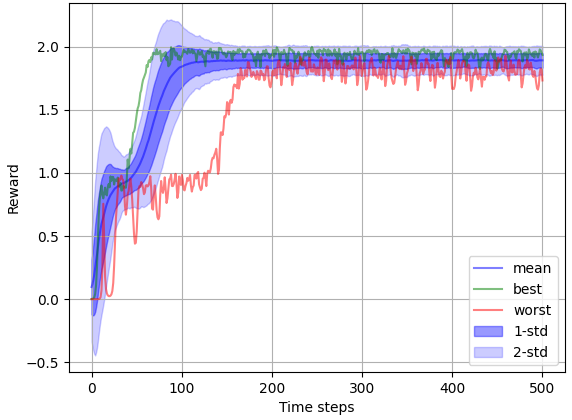}
       \caption{Rewards summary}
   \end{subfigure}
   
   \caption{Evaluation for the ``go to pose'' task, with 1024 parallel agents running for 500 steps (25s), each with randomized initial conditions.}
   \label{fig:gotopose_eval}
\end{figure*}

\subsection{DRL agents}
The agents used to evaluate RANS are leveraging PPO (Proximal Policy Optimization)~\cite{schulman2017proximal} policies with multi-discrete action-space to solve GoToXY and GoToPose tasks in the 3 DoF scenarios. The agents are modeled as single actor-critic networks with two hidden layers with 128 neurons and are trained for 2000 epochs. For the 6~DoF scenarios, the same PPO agents are used, but the networks are larger, with three hidden layers with 256 neurons each, to deal with the increased action and observation spaces. Preliminary results demonstrated that PPO policies effectively solved the designated tasks with high accuracy. 

\section{Experimental Setup and Results}

\subsection{3~DoF Pose Evaluation results}
The evaluation encompassed spawning a trained policy in random poses around the target, within distances of 3 to 4 meters. Evaluation metrics, including distance-to-target over time and equivalent planar trajectories, were utilized to quantify and visualize the performance of the trained agents.
Figure~\ref{fig:gotopose_eval} shows the results of the evaluation of an agent trained for the GoToPose task over 1024 runs under nominal conditions. In $(a)$ the average number of thrusts activation per episode shows a high energy demand, needed to achieve both position and orientation control, which can need constant compensation as there is distinguished actuator for that (e.g. reaction wheels). $(b)$ and $(c)$ illustrate the fast and stable convergence to the target position through the distance lines starting from a random position between 3 and 4 meters. Similarly, the plots $(f)$ and $(g)$ demonstrates the convergence to the target orientation. In $(d)$ the mean and standard deviation of the angular velocity show the rotation ranges, after an initial spike, tend to quickly converge to zero, or oscillate around $0.3$ radians in the worst case. All the trajectories can be seen in the 
2D plane $(e)$, where starting from a random position they all converge to the center. Finally, the reward best, mean and worst cases in $(h)$ interestingly display the agents learned behavior to first collect the orientation rewards by adjusting the attitude, then moving to the target position. 

\begin{figure*}[!htb]
    \begin{tabular}{ccc}
        \includegraphics[width=0.3\linewidth]{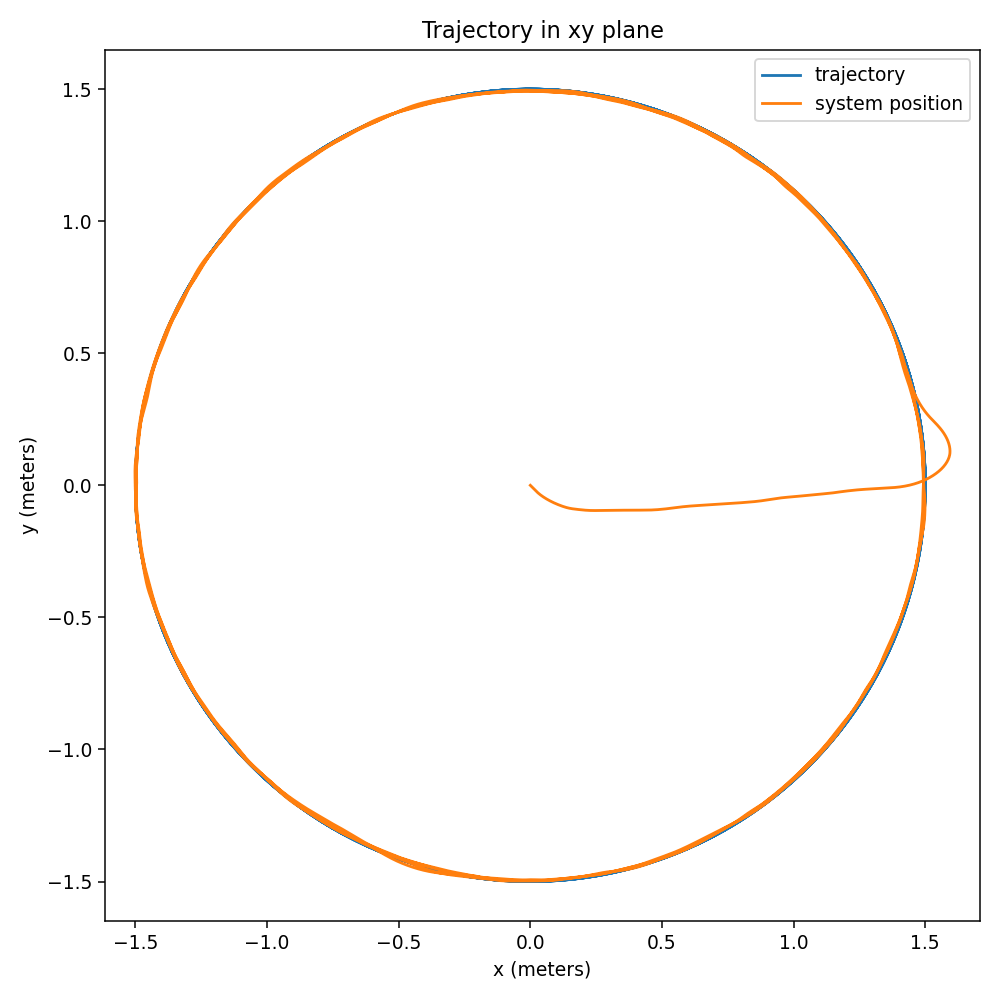} & \includegraphics[width=0.3\linewidth]{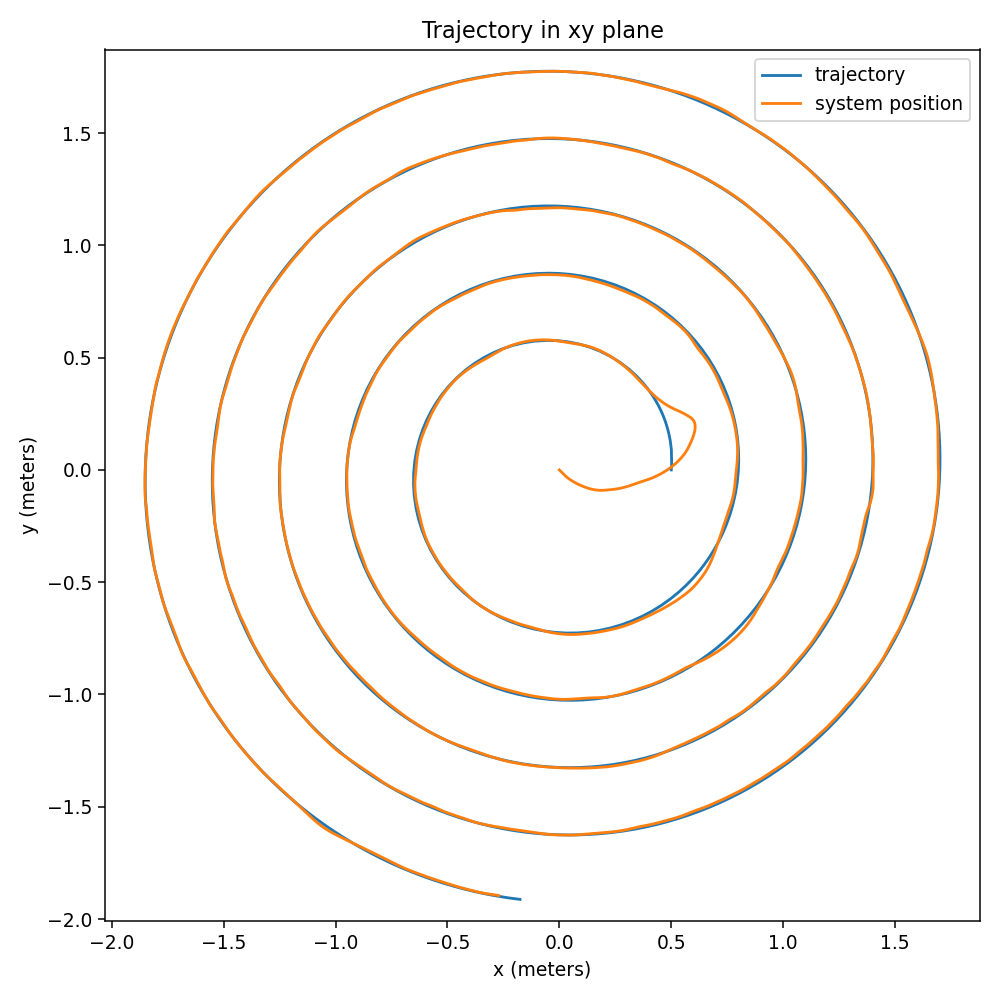} & \includegraphics[width=0.3\linewidth]{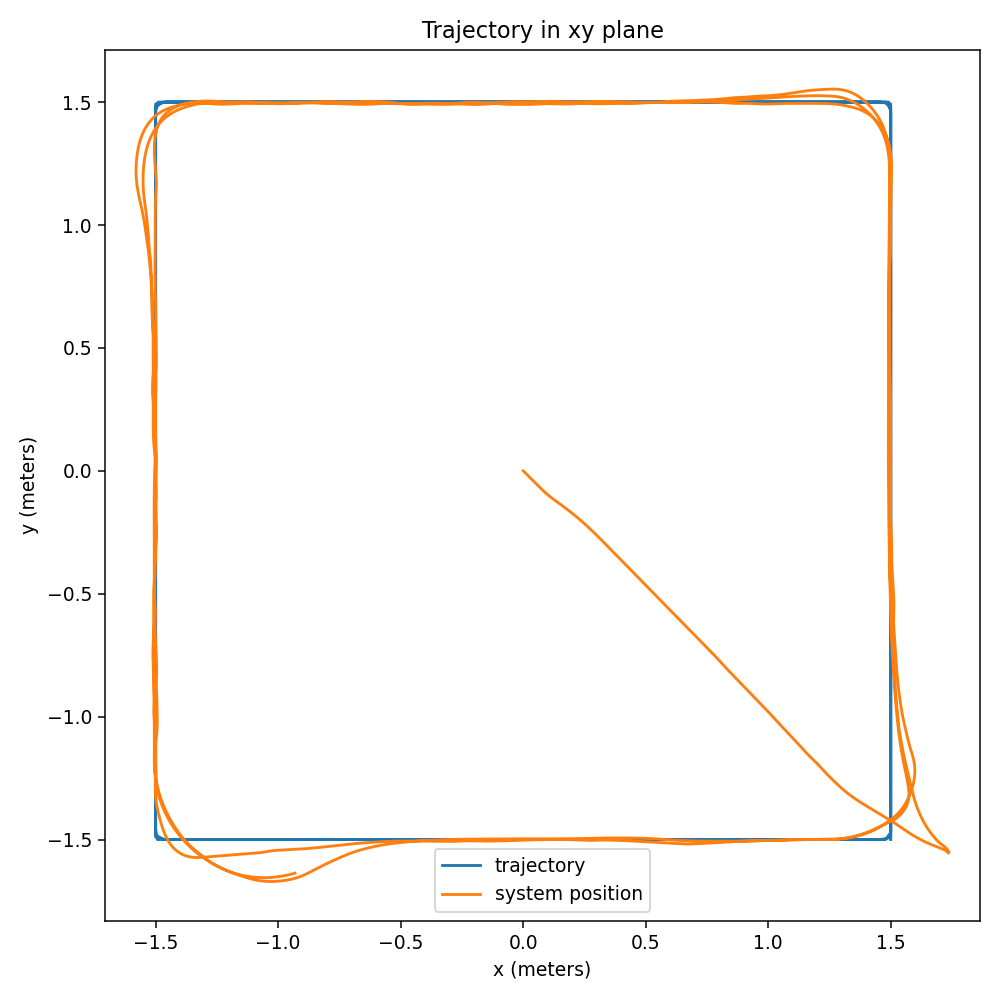}\\
        \includegraphics[width=0.3\linewidth]{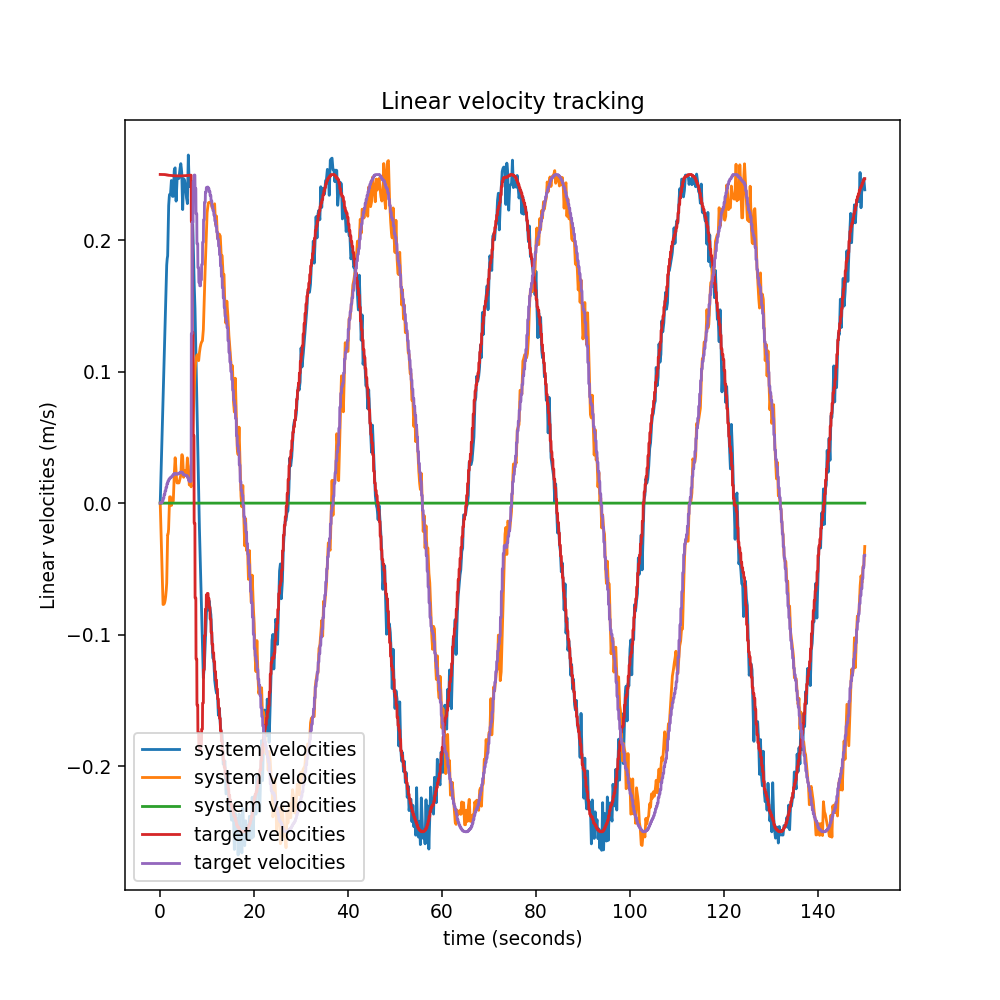} & \includegraphics[width=0.3\linewidth]{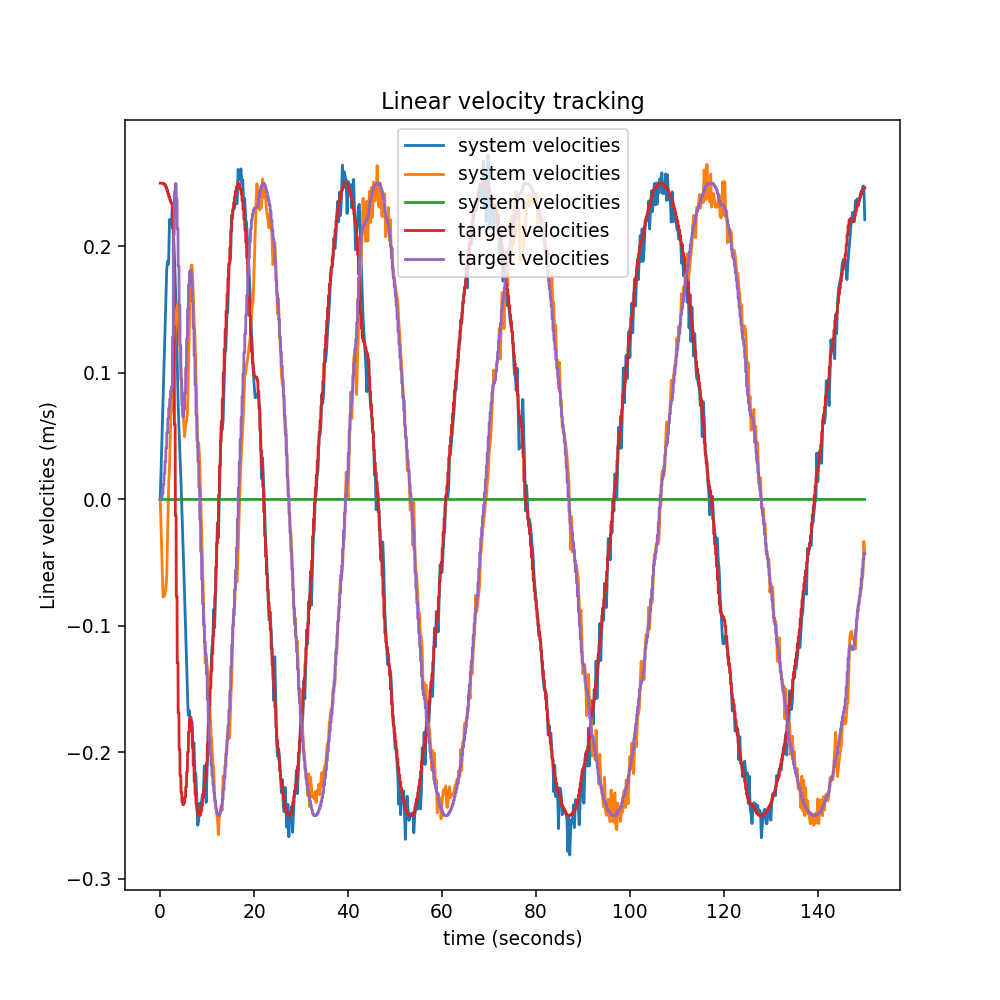} & \includegraphics[width=0.3\linewidth]{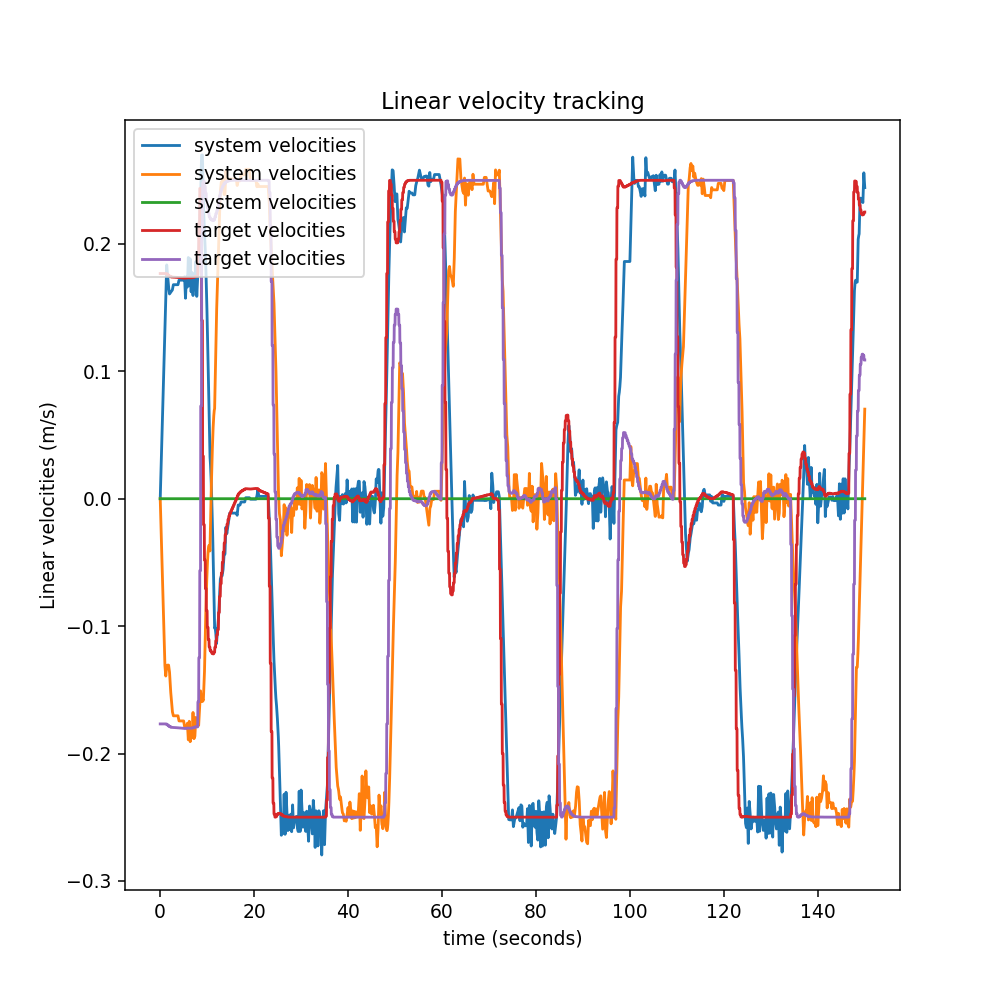}\\
         & 
    \end{tabular}
    \centering
    
    \caption{Example of trajectory following using the agents trained to track velocities. From left to right, circle, spiral, and square trajectories are followed using a simple look-ahead controller that provides the RL agents with target velocities.}
    \label{fig:velocity_tracking}
\end{figure*}

\begin{figure*}
    \centering
    \includegraphics[width=\linewidth]{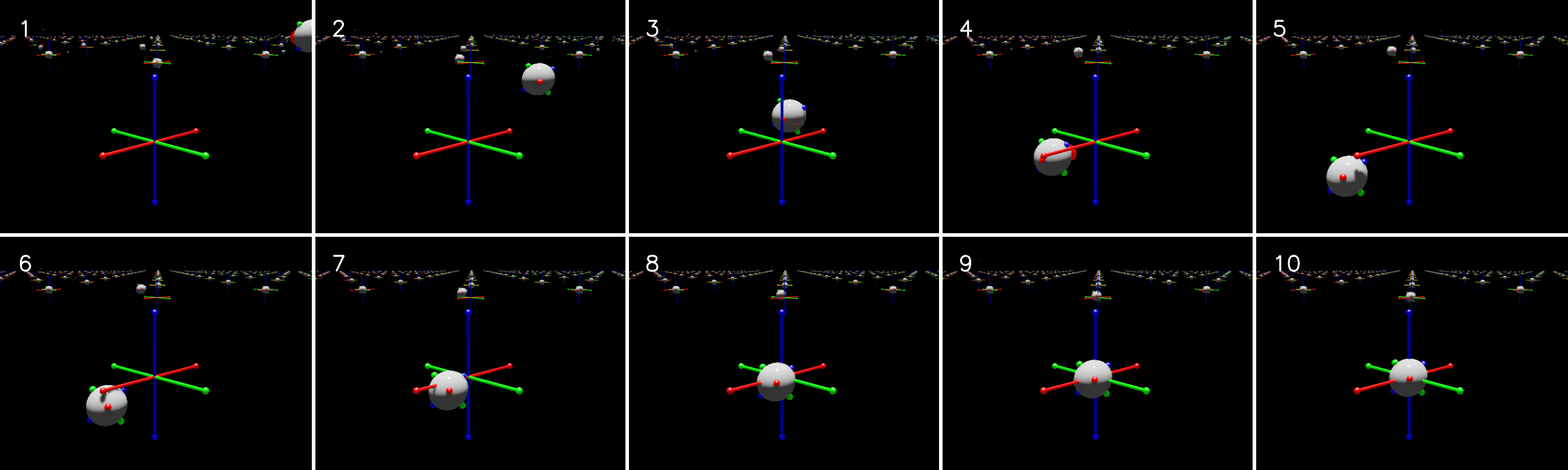}
    \caption{Visualization of a 10-frame sequence extracted from a rendered episode depicting the 6-DoF GoToXYZ task. Progressing from the top left to the bottom right, this sequence offers a detailed view of one of the 1024 agents approaching and effectively stabilizing at the designated target location.}
    \label{fig:6dof_episode}
\end{figure*}

\subsection{3~DoF Linear Velocity Tracker results}

The results of the linear velocity tracking policy trained with PPO are demonstrated through a set of trajectory following examples.
To make the agent follow the trajectories, we use a simple look-ahead planner.
In these examples, showed in Fig.~\ref{fig:velocity_tracking}, the RL agents exhibit the ability to accurately track target velocities, enabling them to follow both simple and more complicated predefined trajectories with precision.
This planner acquires the farthest point of the trajectory within a 25cm radius, or if there are no points within 25cm, the closest point to the system.
Using the position of this point and the position of the system, we compute vector between these two points, normalize it, multiply it by the desired system velocity (0.25m/s) and the resulting vector is given to the agent as the velocity to be tracked.
Three distinct trajectories are showcased, including circle, a spiral, and a square.
On the circle trajectory, we can see that the agent can easily track sinusoidal velocity commands, though the measured velocities are noisy.
Similarly, the spiral shows the agent successfully tracking sinusoidal velocities with different frequencies.
The most challenging trajectory to track is the square, this results in step-like velocity commands which the agents match fairly well.
However, we can see that the positive and negative velocities have less overshoot than the null velocities.
This could be linked to the reward design.

\subsection{6~DoF GoToXYZ Evaluation results}

Presented here is an illustrative demonstration of the policy's behavior trained for the 6DoF GoToXYZ task. The agent's initialization occurs on a spherical surface centered around the target, randomly positioned within a radius of 1 to 5 meters and an angle $\phi$ ranging from $-\pi$ to $\pi$. During the evaluation episode, a set of 1024 parallel agents is spawned for the task, converging swiftly towards the target. Occasionally, some agents display small overshooting or high angular speeds. Overall, the PPO agent's performance is acceptable, exemplified by a rendered trajectory illustrated in Figure\ref{fig:6dof_episode}.

\section{Discussion and Future Directions}
The results obtained from the experiments validate the effectiveness of RANS in enabling RL-based control policies for spacecraft navigation. The high accuracy achieved in solving navigation tasks underscores the potential of RANS in advancing research in the field of aerospace robotics. However, it is essential to acknowledge the preliminary nature of these results and recognize the need for extensive evaluation across a broader spectrum of scenarios to validate the robustness and versatility of RANS.
By extending the capabilities of the Isaac Gym framework, RANS offers a discrete force-based module designed explicitly to train RL agents for thrust-based spacecraft control. The open-source nature of RANS fosters collaboration and innovation within the aerospace and robotics communities, enabling researchers to exploit parallelized simulation frameworks for spacecraft navigation.

Future developments in RANS will focus on expanding the available tasks to encompass velocity and orientation tracking, further enhancing the realism of simulations. Additionally, customization features, such as reconfigurable thruster settings and failure scenarios, will be integrated to offer a more comprehensive simulation environment. The software for RANS is openly accessible at \url{https://github.com/elharirymatteo/RANS/tree/ASTRA23}, encouraging widespread utilization and continued enhancements in line with evolving research needs.

% The following bibliography was produced with
%   \bibliographystyle{aa}
%   \bibliography{esapub}
% The results are inserted directly here to simplify
% the demonstration.

\bibliography{main}
%\begin{thebibliography}{}
%\bibitem[Allen(1973)]{allen73}
%Allen C., 1973, Astrophysical Quantities, Athlone Press

%\bibitem[Nobody et~al.(1997)]{nobody97}
%Nobody B., Somebody G., Who M.E., et~al., 1997, ApJ 331, 902

%\bibitem[Smith \& Jones(1996)]{smith96}
%Smith A., Jones B., 1996, A\&A 555, 999

%\end{thebibliography}
\end{document}